\pdfoutput=1

\documentclass[11pt]{article}

\usepackage[]{acl}
\PassOptionsToPackage{table}{xcolor}

\usepackage{times}
\usepackage{latexsym}

\usepackage[T1]{fontenc}

\usepackage[utf8]{inputenc}

\usepackage{microtype}

\usepackage{inconsolata}

\usepackage{times}
\usepackage{latexsym}

\usepackage[normalem]{ulem}
\usepackage{mathrsfs}
\usepackage{amsthm}
\usepackage{amsmath}
\usepackage{amssymb, bm}
\usepackage{multirow}
\usepackage{booktabs}
\usepackage{enumitem}
\usepackage{times}
\usepackage{latexsym}
\usepackage{array}
\usepackage{multirow}
\usepackage{graphicx}
\usepackage{amsfonts}
\usepackage{subcaption}
\usepackage{xspace}
\usepackage{tabularx}
\usepackage{algorithm}
\usepackage{arydshln}
\usepackage{algpseudocode}
\usepackage{colortbl}
\usepackage{tcolorbox}
\usepackage{makecell}

\usepackage{longtable}
\usepackage{color, colortbl}

\usepackage{graphicx}
\usepackage{arydshln}
\usepackage{booktabs}
\usepackage{multirow}
\usepackage{caption}
\usepackage{subcaption}
\usepackage{makecell}
\usepackage{csquotes}
\usepackage{epigraph}
\usepackage{wrapfig}
\usepackage{import}
\usepackage{chemformula} 

\definecolor{mypurple}{RGB}{111,61,121}
\definecolor{myblue}{RGB}{46,88,180}
\definecolor{myred}{RGB}{181,68,106}
\definecolor{textorange}{RGB}{237,125,49}
\definecolor{textblue}{RGB}{46,117,181}
\definecolor{textgreen}{RGB}{112,173,71}

\newcommand{\ours}{MetaEOL\xspace}
\title{
Meta-Task Prompting Elicits Embeddings from Large Language Models
}

\author{Yibin Lei\textsuperscript{1}\thanks{Corresponding to: Yibin Lei (e-mail: y.lei@uva.nl) and Chongyang Tao (e-mail: chotao@microsoft.com).}, Di Wu\textsuperscript{1}, Tianyi Zhou\textsuperscript{2}, Tao Shen\textsuperscript{3}, Yu Cao\textsuperscript{4}, \\ \textbf{Chongyang Tao\textsuperscript{5}\footnotemark[1], Andrew Yates\textsuperscript{1}} \\
\textsuperscript{1}{University of Amsterdam} \quad \textsuperscript{2}{University of Maryland} \\
\textsuperscript{3}{AAII, FEIT, University of Technology Sydney}
\quad \textsuperscript{4}{Tencent IEG} \quad \textsuperscript{5}{Microsoft Corporation}\\
\texttt{\{y.lei, d.wu, a.c.yates\}@uva.nl}, 
\texttt{tianyi@umd.edu} \\
\texttt{tao.shen@uts.edu.au},
\texttt{rainyucao@tencent.com},
\texttt{chotao@microsoft.com}
}

\begin{document}
\maketitle
\begin{abstract}
We introduce a new unsupervised text embedding method, Meta-Task Prompting with Explicit One-Word Limitation (\ours), for generating high-quality sentence embeddings from Large Language Models (LLMs) without the need for model fine-tuning. Leveraging meta-task prompting, \ours guides LLMs to produce embeddings through a series of carefully designed prompts that address multiple representational aspects. Our comprehensive experiments demonstrate that embeddings averaged from various meta-tasks are versatile embeddings that yield competitive performance on Semantic Textual Similarity (STS) benchmarks and excel in downstream tasks, surpassing contrastive-trained models. Our findings suggest a new scaling law, offering a versatile and resource-efficient approach for embedding generation across diverse scenarios.\footnote{Our code is publicly available at~\url{https://github.com/Yibin-Lei/MetaEOL}.}
\end{abstract}

\section{Introduction}
The advent of Large Language Models (LLMs) such as GPT-3~\citep{brown2020language} and LLaMA~\citep{touvron2023llama} has marked a significant milestone in the field of natural language processing (NLP), introducing promising unsupervised methods for various NLP tasks by leveraging task-related instructions or prompts~\citep{qin-etal-2023-chatgpt, zhong2023chatgpt, zhao2023gptbias}. 
These tasks also include the generation of sentence embeddings, which aims to produce sentence representations that can be applied across a wide range of scenarios. They have been applied to intrinsic tasks like Semantic Textual Similarity (STS)~\citep{agirre-etal-2012-semeval,cer-etal-2017-semeval}, to downstream tasks including information retrieval~\citep{mitra2017learning,izacard2021unsupervised}, and to sentiment classification~\citep{ke-etal-2020-sentilare} and beyond. 
By employing specific prompts \citep{jiang2023scaling, jiang2022promptbert}, researchers have begun to explore the potential of extracting meaningful sentence embeddings directly from the hidden states of LLMs without the need for explicit training. 
These prompt-based approaches generate embeddings without the need for any fine-tuning or in-context learning, which is a substantial improvement over approaches that require extensive fine-tuning to achieve high performance.

\begin{figure}[t]
    \centering
    \includegraphics[width=0.40\textwidth]{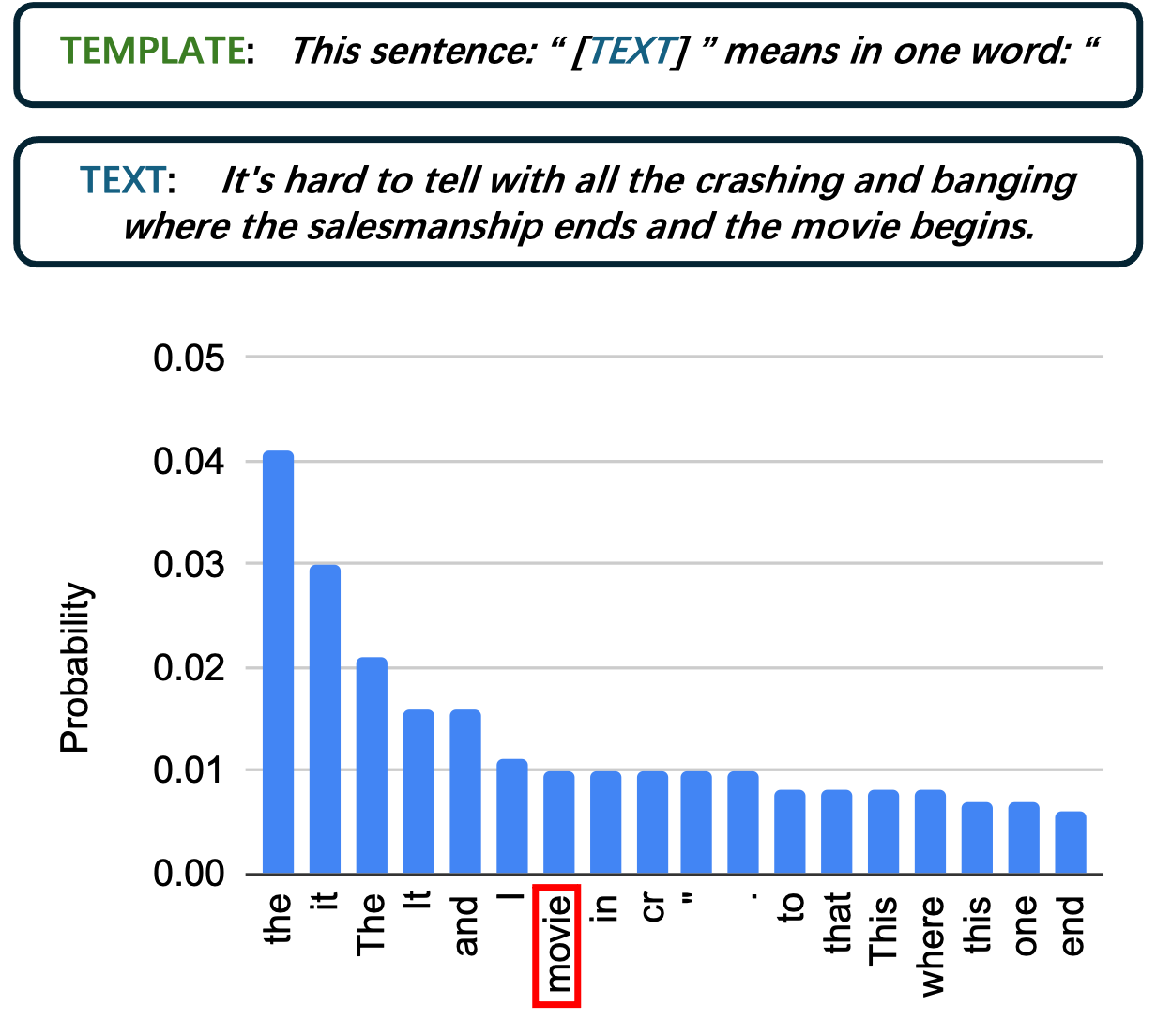}
    \caption{The highest decoding probabilities are largely allocated to stop words that carry little useful information when conducting a meaning compression prompting, even if employing a constraint of "\textit{in one word}" following~\citet{jiang2023scaling}. Although the general semantic, \textcolor[RGB]{235,50,35}{movie}, is contained, other aspects of this sentence are missing, like sentiments.}
    \label{fig:fig-1}
\end{figure}

Initial efforts in this domain, as highlighted by works like~\citep{jiang2023scaling, jiang2022promptbert, meaning_representation}, have focused on unsupervised techniques that extract sentence representations directly from LLMs. These methods typically involve using fill-in-the-blanks prompts, such as \textit{This sentence: "[TEXT]" means in one word:"}~\citep{jiang2023scaling}, to embed a sentence into a single token representation by using the output hidden state of the last token as the sentence's embedding.
While they perform well, these approaches also reveal the inherent challenges of this task: embeddings may be overly simplistic or misaligned with the intended semantic nuances of the sentences. 

In a pilot experiment illustrated in Figure~\ref{fig:fig-1}, we demonstrate that a previous prompt-based method~\citep{jiang2023scaling} can struggle to capture a sentence's meaning, especially when the usage of the sentence is associated with multiple aspects. When probing the probability distribution for the next token during decoding, which reflects the embedding quality of the last token\footnote{The decoding probabilities are derived by comparing the similarity between the output hidden state of the last token and the token embeddings of the whole vocabulary.}, the highest probabilities are mostly distributed to frequent stop words. Although the general \textit{movie} topic appears, other meaningful aspects like sentiments are missing.

A straightforward solution to mitigate this issue is to provide LLMs with task-specific instructions. This approach involves instructing the model with prompts explicitly designed for a particular task, thereby tailoring the embeddings to better suit the specific requirements of that task. However, 
considering the wide range of distinct tasks that an embedding may be used for~\citep{ni, wang-etal-2022-superni, chung2022flanv2}, 
this would be impractical.
Furthermore, while task-specific embeddings are effective for their corresponding tasks, they may fail to generalize well across different tasks.

Inspired by the principles of the usage-based theory of language acquisition~\citep{tomasello2009usage}, which asserts that the essence of meaning is rooted in the practical utilization of language, 
our approach aims to generate broad embeddings through the use of meta-task prompting,
inspired by meta-task prompted training~\citep{sanh2022multitask} and hyper-prompt~\citep{pmlr-v162-hyperprompt} techniques. By defining a suite of meta-tasks, each tailored to a distinct application context, \ours prompts LLMs to consider multiple representational tokens from a variety of perspectives. This multifaceted approach enables the extraction of more diverse and nuanced contextualized token embeddings that collectively form a comprehensive sentence embedding.

Extensive experiments empirically show that: (i) Simply averaging embeddings from different meta-tasks without any training leads to general embeddings that are competitive to contrastive-trained models on STS tasks and can achieve the best average result on several downstream tasks.
(ii) Incrementally integrating more meta-tasks (ranging from one to four) yields consistent improvements across STS tasks, showcasing high generalities, and highlighting the significant impact of meta-task integration on overall performance.
(iii) The final layer is not always the most effective for STS tasks and with a simple proportional layer selection strategy, we achieve the best results with a 70B model, which points to a potential scaling law.

\begin{figure*}[tb]
    \centering
    \includegraphics[width=0.99\textwidth]{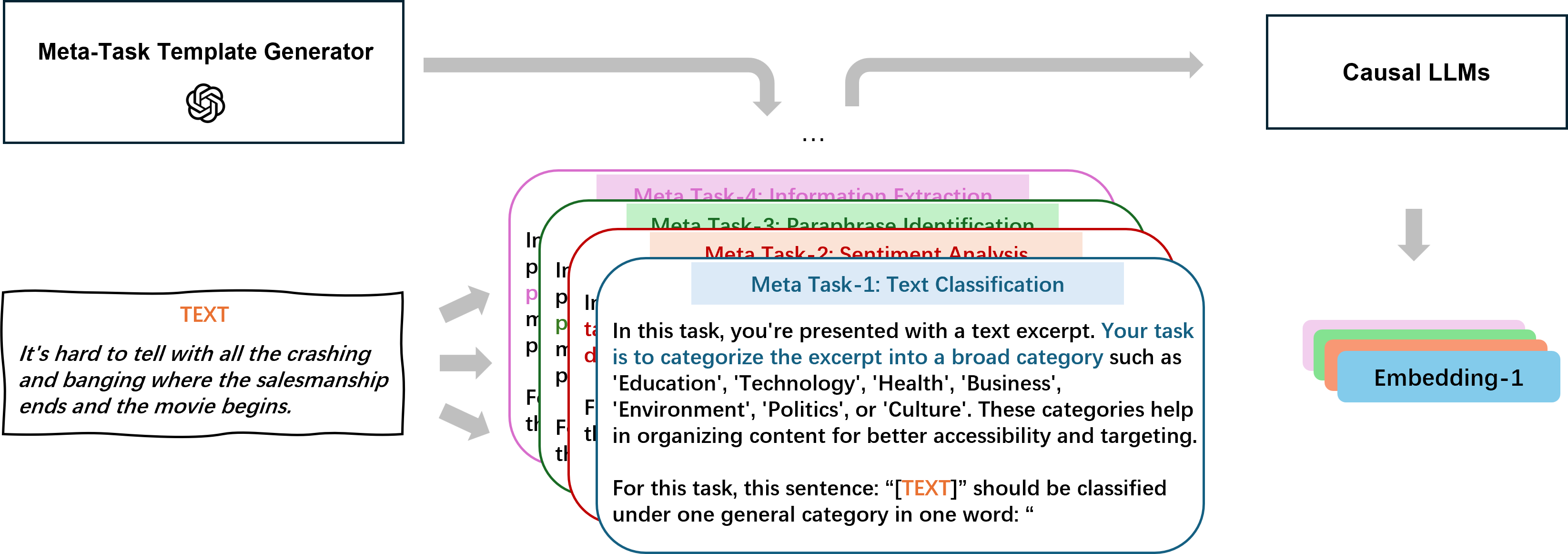}
    \caption{The workflow of our method (MetaEOL). We use the prompt in Appendix~\ref{sec:template_generate} to prompt ChatGPT-4 to generate templates. Each input sentence will be decorated with multiple task-specific templates, indicating its various intended usage scenarios. The resulting multiple prompts will be fed to LLMs. Then, multiple task-specific embeddings will be extracted. The final sentence embedding is obtained by averaging the task-specific embeddings.}
    \label{fig:workflow}
\end{figure*}

\section{Related Work}
\paragraph{Sentence Embeddings.}
Sentence embeddings aim to encode the semantic content of sentences into fixed-sized vector representations. 
Recent developments in contrastive learning have proven to be highly effective for generating sentence embeddings, under both unsupervised and supervised settings~\citep{gao2021simcse, jiang2022promptbert, chuang-etal-2022-diffcse, wu-etal-2022-pcl}. 
For instance, SimCSE~\citep{gao2021simcse} utilizes different dropout masks as a form of noise to create positive pairs in an unsupervised fashion, while in a supervised setting, models like Sentence-BERT~\cite{reimers-gurevych-2019-sentence} leverage natural language inference (NLI) datasets to construct positive and negative pairs. Additionally, \citet{su-etal-2023-one} and \citet{asai-etal-2023-task} show that training with a large amount of tasks with annotated instructions can enable the model to generate embeddings tailored to different downstream tasks. In contrast, our approach \ours demonstrates the potential of utilizing LLMs directly to generate instruction-followed embeddings without the need for any additional training.

\paragraph{Large Language Models for Text Representation.}
Recent studies have explored the use of LLMs for enhancing text embeddings through data augmentation techniques~\citep{cheng-etal-2023-improving-feedback, zhang-etal-2023-contrastive-learning_scratch}. Notably, Sentence-T5~\cite{ni-etal-2022-sentence} and GTR~\cite{ni-etal-2022-gtr} employ contrastive learning on models with billions of parameters. More recently, research has focused on converting an LLM directly into a text encoder without any training. \citet{liu2023meaning} represents sentences through the distribution of possible text continuations, comparing the distributional similarity between sentences. This method, although effective, necessitates the generation of 20 trajectories, each up to 20 tokens in length, making it computationally intensive. 
\citet{jiang2022promptbert} incorporates in-context learning~\citep{icl_survey} to enhance sentence embeddings. While proven effective, it also reveals that the produced embeddings are task-specific and struggle with generalization across various downstream tasks, in addition to being highly sensitive to the choice of demonstrations. Contemporaneous works investigate the potential of LLMs either by repeating the input texts~\cite{springer2024repetition} or enabling bidirectional attention~\cite{behnamghader2024llm2vec} to address the lack of backward dependency in LLMs.
Additionally, the potential of LLMs under supervised training settings has also been studied in recent works~\cite{ma2023repllama, li2024bellm, li2024angleoptimized, wang2024e5mistral, li2024ese, lee2024nvembed}.

\paragraph{Multitask Prompts.}
Studies have demonstrated that models fine-tuned using multi-task prompts and datasets can serve as general-purpose models with strong capabilities in generalizing to new tasks~\citep{sanh2022multitask, wei2022flan, chung2022flanv2, wang-etal-2022-superni, ni}. Our approach \ours aligns with this concept, showcasing that multi meta-task prompts can similarly generate general-purpose embeddings, remarkably without necessitating any training.

\section{Method}
In this section, we begin by reviewing two kinds of previous prompting methods for deriving sentence representation from masked and causal language models, respectively (Section~\ref{sec:former_method}). Subsequently, we describe our proposed method, i.e., \textbf{Meta}-Task Prompting with \textbf{E}xplicit \textbf{O}ne-Word \textbf{L}imitation (MetaEOL) in detail (Section~\ref{sec:method_prompting}). Lastly, we describe meta-tasks involved in this paper (Section~\ref{sec:method_meta_tasks}).

\subsection{Previous Language Model Prompting}\label{sec:former_method}
\subsubsection{Masked Language Model}\label{sec:MLM}
Masked language models, e.g., BERT~\citep{devlin-etal-2019-bert} and RoBERTa~\citep{liu2019roberta}, use a mask prediction task to capture contextual information for a certain token. To align with this point, PromptBERT~\citep{jiang2022promptbert} formulates the sentence embedding extraction as a similar task and employs the following template,

\begin{center}
\textit{This sentence : “[TEXT]” means [MASK] .} 
\end{center}

\noindent for prompting. Here, [TEXT] and \textit{[MASK]} indicate the placeholder for the input sentence and the mask token. The last layer's hidden vector of \textit{[MASK]} token is directly used as the sentence representation. 

\citet{jiang2022promptbert} empirically show that such a simple prompting method can achieve decent performance, and equipping it with a contrastive loss for large-scale continued training leads to further enhancements for embedding quality. However, it is worth noting that extra training is resource-intensive, especially for today's LLMs. To enhance clarity, we provide results both with and without training on BERT and RoBERTa in the following experiments.

\subsubsection{Causal Language Model}\label{sec:CLM}
Others have investigated directly extracting sentence representation from large Causal Language Models (CLMs), e.g., OPT~\cite{zhang2022opt} or LLAMA~\cite{touvron2023llama}, without additional training. Inspired by ~\citet{jiang2022promptbert}, PromptEOL~\citep{jiang2023scaling} employs a similar prompting template as follows,

\begin{center}
\textit{This sentence: “ [TEXT] ” means in one word:“} 
\end{center}

\noindent where the last layer's hidden vector for the last token ``\textit{“}'' is extracted as the sentence representation. A constraint of ``in one word'' is applied to avoid the model's tendency to generate long sentences such that the last token fails to capture the overall information. 

However, the obtained embedding highly relies on the single prompt, which confines the inference process and can result in non-comprehensive features. For example, as shown in Figure~\ref{fig:fig-1}, for a negative review of a movie, the resulting embedding does not capture critical aspects such as sentiment.

\subsection{Meta-Task Prompting}\label{sec:method_prompting}
To overcome the issues raised above, we propose \textbf{Meta}-Task Prompting with \textbf{E}xplicit \textbf{O}ne-Word \textbf{L}imitation (MetaEOL). 
A meta-task is associated with a potential broad usage scenario for the corresponding sentence representation. As shown in Figure~\ref{fig:workflow}, we directly prompt casual LLMs with the goals of multiple meta tasks, aiming to obtain the representations under various broad intents.

Specifically, we produce task-oriented prompts by decorating the original prompting template used for causal LLMs (Section~\ref{sec:CLM}) with the corresponding task description. For example, given a meta-task where representations are extracted for Text Classification (TC), we extend the template with task-oriented context to define the behavior during inference. 
As shown in the template of \textit{Meta Task-1} in Figure~\ref{fig:workflow}, a detailed task description text is placed at the beginning of the prompt, instructing the LLM to categorize the excerpt into a broad category.
Then, an instruction with a constraint of ``in one word'' is followed to ensure models aggregate the information of the whole sentence into the embedding of the last token. The placeholder \textit{[TEXT]} will be substituted with the original sentence to produce the final task-oriented prompt.
The resulting task-specific prompt will serve as input to LLMs. Subsequently, we extract the hidden vector of the last token ``\textit{“}'' as the sentence representation, following the pattern outlined in Section~\ref{sec:former_method}. 

It is worth noting that given various meta-tasks, distinct templates will be employed, leading to multiple different sentence embeddings. Our hypothesis is that each embedding captures a distinct representation customized for a specific feature viewpoint (meta-task). In this paper, we empirically show that simply averaging different embedding derived from multiple meta-tasks can achieve superior performance for both intrinsic and downstream evaluation benchmarks. 

\subsection{Types of Meta-Tasks}\label{sec:method_meta_tasks}
In this paper, we conduct experiments on the following four distinct meta-tasks, i.e., Text Classification (TC), Sentiment Analysis (SA), Paraphrase Identification (PI), and Information Extraction (IE), aiming to capture information from different angles. For example, intuitively, the TC task primarily emphasizes topic-level information, whereas the IE task concentrates on surface-level signals.

For each meta-task, we straightforwardly leverage ChatGPT-4 as a template generator to produce multiple templates. The instruction we used to prompt the ChatGPT-4 is provided in Appendix~\ref{sec:template_generate}. 

Note that introducing more meta-tasks is trivial, because it only requires adding more task names to the generator. Here, we choose the above four meta-tasks as a testbed to assess scalability. More specifically, in Section~\ref{sec:number_of_tasks}, we show that incrementally adding more meta-tasks to our workflow results in consistently better performance.

\section{Experiments}
\begin{table*}[th]
\centering
\normalsize
\setlength{\tabcolsep}{5pt}
\resizebox{\linewidth}{!}{%
\begin{tabular}{lrccccccccc}
\toprule
\textbf{Method} & \textbf{Params} & \textbf{STS12} & \textbf{STS13} & \textbf{STS14} & \textbf{STS15} & \textbf{STS16} & \textbf{STS-B} & \textbf{SICK-R} & \textbf{Avg.}\\
\midrule
\midrule
\multicolumn{10}{l}{\it{Unsupervised Contrastive Training}}\\
SimCSE-BERT  & 110M & 68.40 & 82.41 & 74.38 & 80.91 & 78.56 &    76.85     &      72.23      & 76.25 \\
SimCSE-RoBERTa & 123M & 70.16 &  81.77 & 73.24 & 81.36 & 80.65 & 80.22 & 68.56 & 76.57 \\
PromptBERT  & 110M & 71.56 & 84.58 & 76.98 & 84.47 & 80.60 &  81.60 & 69.87& 78.54 \\
PromptRoBERTa & 123M &  73.94 & 84.74 & 77.28 & 84.99 & 81.74 & 81.88 & 69.50 & 79.15 \\
{LLM2Vec-LLAMA2}
& 7B & 65.39 & 79.26 & 72.98 & 82.72 & 81.02 & 78.32 & 71.77 & 75.92 \\
{LLM2Vec-Mistral}
& 7B & 67.65 & 83.90 & 76.97 & 83.80 & 81.91 & 80.42 & 75.55 & 78.60 \\
\midrule
\midrule
\multicolumn{10}{l}{\it{Without Contrastive Training}}\\
BERT avg. & 110M & 30.87 & 59.89 & 47.73 & 60.29 & 63.73 & 47.29 & 58.22 & 52.57 \\
BERT prompt & 110M & 60.96  & 73.83 & 62.18 & 71.54 & 68.68 & 70.60 & 67.16 & 67.85 \\
ST5-Enc avg. & 4.8B & 34.97 & 60.19 & 47.59 & 66.40 & 70.62 & 62.83 & 63.57 & 58.02 \\
LLAMA2 avg. & 7B & 35.49 & 53.15 & 40.12 & 55.35 & 53.26 & 42.10 &      49.96 & 47.06\\
Mistral avg. & 7B & 41.13 & 54.08 & 43.99 & 56.94 & 53.80 & 42.99 &      52.32 & 49.32\\
\midrule
{Echo-LLAMA2}
& 7B & 52.40 & 72.40 & 61.24 & 72.67 & 73.51 & 65.73 & 64.39 & 66.05 \\
{Echo-LLAMA2}
& 13B & 59.36 & 79.01 & 69.75 & 79.86 & 76.75 & 71.31 & 70.27 & 72.33 \\
{PromptEOL-LLAMA2}
& 7B & 58.81 & 77.01 & 66.34 & 73.22 & 73.56 & 71.66 & 69.64 & 70.03 \\
{PromptEOL-Mistral}
& 7B & 63.08 & 78.58 & 69.40 & 77.92 & 79.01 & 75.77 &	69.47 & 73.32 \\
{PromptEOL-LLAMA3}
& 8B & 60.88 & 78.57 & 68.18 & 76.75 & 77.16 &    72.83     &      68.94      & 71.90 \\
{PromptEOL-LLAMA2}
& 13B & 56.19 & 76.42 & 65.42 & 72.73 & 75.21 & 67.96 & 68.23 & 68.83\\
\midrule
{\ours-LLAMA2 (\textbf{\textit{Ours}})}
& 7B & 64.16 & 81.61 & 73.09 & 81.11 & 78.94 & 77.96 & 74.86 & 75.96 (+5.93)\\
{\ours-Mistral (\textbf{\textit{Ours}})}
& 7B & 64.05 & 82.35 & 71.57 & 81.36 & 79.85 & 78.29 & 75.13 & 76.09 (+2.77)\\
{\ours-LLAMA3 (\textbf{\textit{Ours}})}
& 8B & 65.10 & 83.08 & 73.01 & 81.87 & 81.47 & 80.47 & 76.46 & 77.35 (+5.45)\\
{\ours-LLAMA2 (\textbf{\textit{Ours}})}
& 13B & 61.07 & 82.53 & 73.30 & 80.99 & 79.14 & 77.11 & 74.77 & 75.56 (+6.73)\\
\bottomrule
\end{tabular}
}
\caption{Results on STS tasks (Spearman correlation scaled by 100x). Values in parentheses, such as ``(+5.93)'' in \ours's results, represent the increase in average score compared to the average score of the same model utilizing PromptEOL.
}
\label{tab:sts}
\end{table*}

\subsection{Settings}
\paragraph{Dataset.} Suggested by prior works~\citep{reimers-gurevych-2019-sentence, gao2021simcse, jiang-etal-2022-promptbert} that an important objective of sentence embeddings is to cluster semantically similar sentences, we evaluate \ours on seven semantic textual similarity (STS) datasets, utilizing the SentEval toolkit~\citep{conneau-kiela-2018-senteval}.
The STS datasets consist of STS 2012-2016~\citep{agirre2012semeval, agirre2013sem, agirre2014semeval, agirre2015semeval, agirre2016semeval}, STS-B~\citep{sts-b2017}, and SICK-R~\cite{marelli2014sick}. Each sentence pair in the STS datasets is annotated with a score from 0 to 5 indicating the pairwise semantic similarity. The Spearman correlation (scaled by 100x) between the model-predicted similarity scores and human-annotated similarity scores is used as the metric. We employ cosine similarity to measure the similarity between sentence embeddings. The Spearman correlation is computed under the ``all'' setting.

\paragraph{Baselines.}
The baselines we consider can be categorized into two types -- models with contrastive training and without contrastive training:
(i) \emph{Models with Contrastive Training:}
We compare \ours with SOTA unsupervised contrastive-trained models, namely SimCSE~\citep{gao2021simcse} and PromptBERT~\cite{jiang2022promptbert}. The models are trained on $10^6$ sentences randomly sampled from Wikipedia. Results based on BERT~\citep{devlin-etal-2019-bert} and RoBERTa~\cite{liu2019roberta} models are reported. Contemporaneous LLM-based approach LLM2Vec~\cite{behnamghader2024llm2vec} is also included for comparison. LLM2Vec comprises three stages: bidirectional attention enabling, masked next token prediction training, and unsupervised contrastive training (similar to SimCSE) to transform an LLM into a text encoder.
Considering (ii) \emph{Models without Contrastive Training:}
We compare \ours with (1) average pooling methods, where average pooling is applied to the output hidden states of all tokens in a sentence to obtain the sentence embedding. We report results with BERT, the encoder of ST5~\citep{ni-etal-2022-sentence}, LLAMA2~\citep{touvron2023llama2} and Mistral~\citep{jiang2023mistral} models; and (2) Prompt-based methods, which include BERT Prompt that employs the same prompt strategy as PromptBERT but does not incorporate contrastive training, PromptEOL and the contemporaneous Echo embeddings~\cite{springer2024repetition}. Echo embeddings repeat the input once and extract embeddings from the second occurrence. 
All methods mentioned above rely on the output from the final layer to obtain the sentence embedding.
\begin{table*}[t]
\centering
\small
\setlength{\tabcolsep}{1pt}
\setlength{\belowcaptionskip}{-0.3cm}
\resizebox{\linewidth}{!}
{%
\begin{tabular}{lll}
\toprule
\textbf{Sentence} & \textbf{Prompt} & \textbf{Top-predicted tokens}\\
\midrule
\multirowcell{5}{Smart and alert, thirteen conversations\\about one thing is a small gem.}
& PromptEOL & {I one a thing the This The smart It it}\\
& Text Classification & {Culture E Pol \textbackslash n Bus " Culture educ Te Health}\\
& Sentiment Analysis & {positive pos good ext good very neut negative smart extremely} \\
& Paraphrase Identification & {smart a the intelligent The short clever conc A conversation} \\
& Information Extraction & {gem smart thing alert small conversation Gem thirteen gem a}  \\
\bottomrule
\end{tabular}}
\caption{The top-10 tokens predicted by different task prompts with Mistral-7B. PromptEOL creates sentence embeddings with an emphasis on stop-word tokens. Text Classification focuses embeddings on topic-relevant tokens like \textit{Culture}. Sentiment Analysis aligns embeddings with sentiment words. Paraphrase Identification diversifies embeddings with synonyms, adding richness with terms like \textit{intelligent}, \textit{short}, and \textit{clever}. Information Extraction steers embeddings toward key factual tokens.}\label{tab:qualitative_examples}
\end{table*}

\paragraph{Implementation Details.}
We apply \ours to LLAMA2-7B, LLAMA3-8B, LLAMA2-13B, and Mistral-7B models, using meta-tasks consisting of Text Classification (TC), Sentiment Analysis (SA), Paraphrase Identification (PI), and Information Extraction (IE). These tasks are distinct and collectively consider diverse aspects of a sentence.  For each of these meta-tasks, we utilize GPT-4 to create two unique task prompts,
resulting in a total of eight task prompts.\footnote{The details of these eight task prompts are presented in Appendix~\ref{sec:metaeol_prompts}.} \ours rely on the output from the final layer to obtain the sentence embedding. We simply average the resulting embeddings of task prompts from different meta-tasks to obtain the final embedding.

\subsection{Main Results}
The results of \ours on STS tasks are shown in Table~\ref{tab:sts}, with notable performance by \ours which requires no training.
Among models that do not require training, prompt-based methods exhibit superior results compared to average pooling methods, especially with the LLAMA and Mistral models. Across various models including LLAMA2-7B/13B, LLAMA3-8B, and Mistral-7B, \ours demonstrates competitive performance compared to contrastive-trained models such as SimCSE-BERT and SimCSE-Roberta, albeit with a slight lag behind PromptBERT. Furthermore, \ours significantly outperforms PromptEOL and Echo embeddings across various test models, demonstrating a consistent improvement. Notably, the LLAMA2-13B model using \ours shows an average improvement of 6.73\% over PromptEOL, underscoring the efficacy of \ours. Compared to LLM2Vec which requires two-stage training, \ours is competitive when using LLAMA2-7B, without the need for any training.

\subsection{Qualitative Example}
We further show the top-10 tokens predicted by different task prompts in Table~\ref{tab:qualitative_examples}. 
The example illustrates that PromptEOL creates sentence embeddings focusing on stop-word tokens (such as \textit{a}, \textit{this}, \textit{the}, \textit{it}), which convey minimal information. In contrast, the four meta-tasks of \ours demonstrably shift the behavior of the embeddings, leading to the prediction of tokens that are distinct and imbued with substantive meaning.

Specifically, Text Classification steers the embeddings toward tokens that are indicative of specific topics, such as \textit{Culture}. Sentiment Analysis is inclined to produce embeddings close to sentiment-related words. Paraphrase Identification yields embeddings that capture a spectrum of synonyms, enriching the sentence with varied linguistic expressions like \textit{intelligent}, \textit{short}, and \textit{clever}. Information Extraction modifies the embeddings towards tokens that represent key facts or elements within the sentence.

\section{Analysis}
In this section, we thoroughly analyze \ours using the LLAMA2-7B model.
\subsection{Ablation Study}
\begin{table}[t]
\centering
\footnotesize
\setlength{\tabcolsep}{4pt}
\setlength{\abovecaptionskip}{0.3cm}
\setlength{\belowcaptionskip}{-0.3cm}
\begin{tabular}{l|c}
\toprule
\textbf{Method} & \textbf{STS Avg.}\\
\midrule
PromptEOL & 70.03  \\
\quad \textit{w.} 7 paraphrases & 62.72   \\
\midrule
MetaEOL & 75.96  \\
\quad TC only & 70.92   \\
\quad SA only & 67.06   \\
\quad PI only & 73.03  \\
\quad IE only & 72.06   \\
\quad \textit{w.} embedding concatenation &  74.99   \\
\quad \textit{w.} max pooling & 72.03 \\

\bottomrule
\end{tabular}
\caption{Ablation study on LLAMA2-7B. STS Avg. refers to the average score of the seven STS tasks. TC: Text Classification; SA: Sentiment Analysis; PI: Paraphrase Identification; IE: Information Extraction.}
\label{tab:ablation}
\end{table}

\begin{table*}[th]
\centering
\footnotesize
\setlength{\tabcolsep}{8pt}
\setlength{\abovecaptionskip}{0.3cm}
\setlength{\belowcaptionskip}{-0.3cm}
\resizebox{\linewidth}{!}{%
\begin{tabular}{lccccccccc}
\toprule
\textbf{Meta-Tasks}\quad\quad\quad\quad\quad & \textbf{STS12} & \textbf{STS13} & \textbf{STS14} & \textbf{STS15} & \textbf{STS16} & \textbf{STS-B} & \textbf{SICK-R} & \textbf{Avg.}\\
\midrule
TC & 58.36 & 75.57 & 67.20 & 77.04 & 74.51 & 71.84 & 71.90 & 70.92 \\
TC+SA &  58.89 & 75.56 & 67.35 & 77.60 & 74.90 & 73.58 & 72.48 & 71.48 \\
TC+SA+PI & 63.08 & 80.01 & 71.24 & 80.38 & 78.26 & 77.42 & 75.00 & 75.06\\
TC+SA+PI+IE & 64.16 & 81.61 & 73.09 & 81.11 & 78.94 & 77.96 & 74.86 & 75.96\\
\bottomrule
\end{tabular}}
\caption{Results on increasing number of tasks with LLAMA2-7B. TC: Text Classification; SA: Sentiment Analysis; PI: Paraphrase Identification; IE: Information Extraction.}
\label{tab:number_tasks}
\end{table*}

We evaluate the effectiveness of key components of \ours in Table~\ref{tab:ablation}. 
First, to ensure the improvement observed with \ours is not merely due to involving more prompts, we create seven paraphrased versions of the PromptEOL prompt, resulting in a total of eight prompts.\footnote{The seven paraphrased prompts are presented in Appendix~\ref{sec:prompteol_paraphrases}} We then average the embeddings from these eight prompts to form the final sentence embedding. We find merely duplicating PromptEOL prompts~(\textit{w.} 7 paraphrase) does not improve PromptEOL but results in a significant decline. Additionally, we implement \ours exclusively on each meta-task (TC/SA/PI/IE only). 
We find that tasks requiring a detailed comprehension of sentences (PI and IE) yield superior performance compared to those requiring a broader understanding, even surpassing PromptEOL. \ours, which combines the embeddings from these meta-tasks, outperforms all individual meta-tasks, confirming the complementarity of the meta-tasks and the effectiveness of combining embeddings from diverse meta-tasks. 
We finally find that averaging the embeddings from different meta-tasks yields better results than either concatenating them or max pooling them across each dimension.

\subsection{Influence of Number of Tasks}
\label{sec:number_of_tasks}
We investigate the influence of the number of tasks as presented in Table~\ref{tab:number_tasks}. We find increasing the number of tasks leads to a consistent improvement in performance on average and nearly every individual STS task. This further verifies the complementarity of the meta-tasks and underscores the importance of utilizing various diverse meta-tasks.

\subsection{Influence of Number of Prompts}
\begin{figure}[t]
    \centering
    \setlength{\abovecaptionskip}{0.1cm}
    \setlength{\belowcaptionskip}{-0.3cm}
    \includegraphics[width=0.85\linewidth]{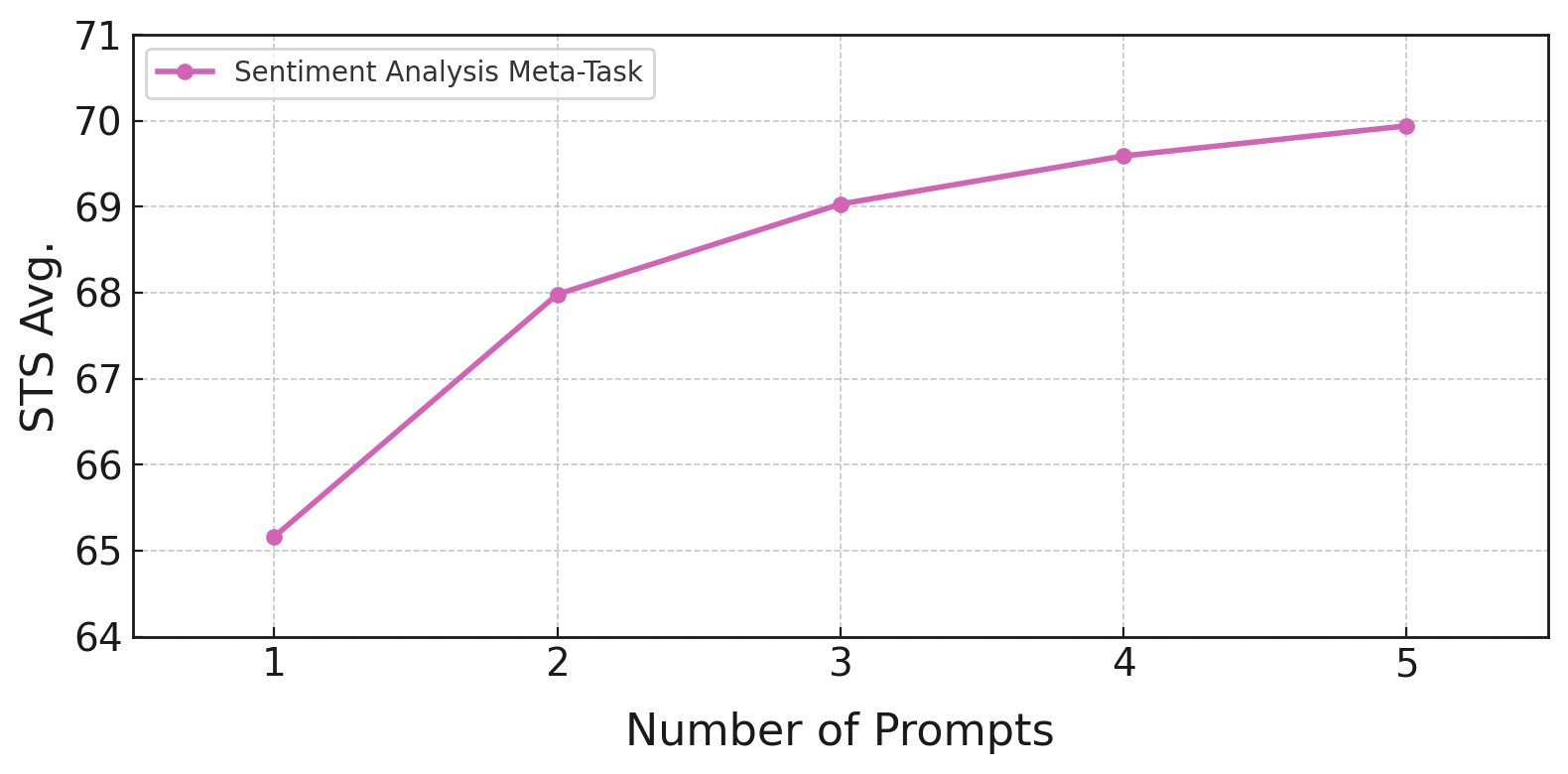}
    \caption{Influence of number of prompts on LLAMA2-7B. STS Avg. refers to the average score of the seven STS tasks.}
    \label{fig:number_instructions}
\end{figure}

Here, we investigate the impact of the number of prompts in Figure~\ref{fig:number_instructions}. 
We concentrate on Sentiment Analysis as the meta-task and utilize GPT-4 to generate three additional Sentiment Analysis prompts besides the two we used in \ours. This results in a total of five distinct prompts, specifically tailored for Product Review Rating, Emotion Detection, Sentiment Polarity Detection, Sentiment Intensity and Emotion Detection, and Aspect-Based Sentiment Analysis, respectively.\footnote{The details of these five instructions are in Appendix~\ref{sec:sa_prompts}.} We systematically computed the average performance across all combinations of these five prompts, conditioned on a fixed number of prompts.

As Figure~\ref{fig:number_instructions} shows, increasing the number of prompts within a particular task type facilitates more nuanced embeddings, thereby leading to better STS results. 
We opt for two prompts for each meta-task for \ours to optimize both performance and computational efficiency.

\subsection{Prompt Sensitivity Analysis}
To test the sensitivity of \ours, we specifically focus on the influence of (i) Tiny perturbations on the task prompt; and (ii) Variations of the major prompting instruction in Appendix~\ref{sec:template_generate}.

\subsubsection{Sensitivity to Tiny Prompt Perturbations}
We apply the synonym replacement operation in~\citet{wei-zou-2019-eda} to replace 10\% of words in the sentiment analysis task prompt of MetaEOL with their synonyms. We craft an additional 4 perturbed prompts. The synonym replacements are sourced directly from WordNet without filtering, which often results in unnatural substitutions, as shown in the Appendix~\ref{sec:tiny_perturbed_prompts}. To provide context for the sensitivity, we included results from SimCSE-BERT-Base with varying random seeds in~\citet{jiang2022promptbert} as a reference.

\begin{table}[ht]
\centering
\small
\setlength{\tabcolsep}{5pt}
\setlength{\belowcaptionskip}{-0.2cm}
{%
\begin{tabular}{lc}
\toprule
\textbf{Method} & \textbf{STS Avg.}\\
\midrule
MetaEOL-LLAMA2-7B  & 67.98±0.67 \\
SimCSE-BERT-Base & 75.42±0.86 \\

\bottomrule
\end{tabular}}
\caption{Sensitivity to tiny prompt perturbations on LLAMA2-7B.}
\label{tab:tiny_perturbations}
\end{table}

The results in Table~\ref{tab:tiny_perturbations} show that even with unnatural substitutions, our MetaEOL-LLAMA2-7B still exhibits a standard deviation of ±0.67 on STS Avg., which is in line with the variance observed in SimCSE-BERT-Base (±0.86), suggesting that our method's sensitivity to prompt perturbations is comparable to that of existing approaches to random seeds.

\subsubsection{Sensitivity to Variations of the Major Prompting Instruction}
We vary the example task prompt in the major prompting instruction in Appendix~\ref{sec:template_generate} with task prompts from the four meta-tasks used in our MetaEOL. As the major prompting instruction is used to prompt ChatGPT-4 to generate task prompts, changing it will lead to a completely different set of task prompts.

\begin{table}[ht]
\centering
\small
\setlength{\tabcolsep}{5pt}
\setlength{\belowcaptionskip}{-0.2cm}
{%
\begin{tabular}{lc}
\toprule
\textbf{Method} & \textbf{STS Avg.}\\
\midrule
MetaEOL-LLAMA2-7B  & 76.17±1.06 \\
SimCSE-BERT-Base & 75.42±0.86 \\

\bottomrule
\end{tabular}}
\caption{Sensitivity to variations of the major prompting instruction on LLAMA2-7B.}
\label{tab:major_variations}
\end{table}

Overall the results show \ours with LLAMA2-7B can beat tuned SimCSE-BERT-Base on average, and is worse but still comparable to SimCSE-Bert-Base in terms of standard deviation, suggesting that MetaEOL's sensitivity to major prompting instruction's variations is also comparable to that of existing approaches to random seeds.

\subsection{Influence of Output Layers}
\label{sec:output_layer}
\begin{figure}[t]
    \centering
    \setlength{\abovecaptionskip}{0.1cm}
    \setlength{\belowcaptionskip}{-0.1cm}
    \includegraphics[width=0.85\linewidth]{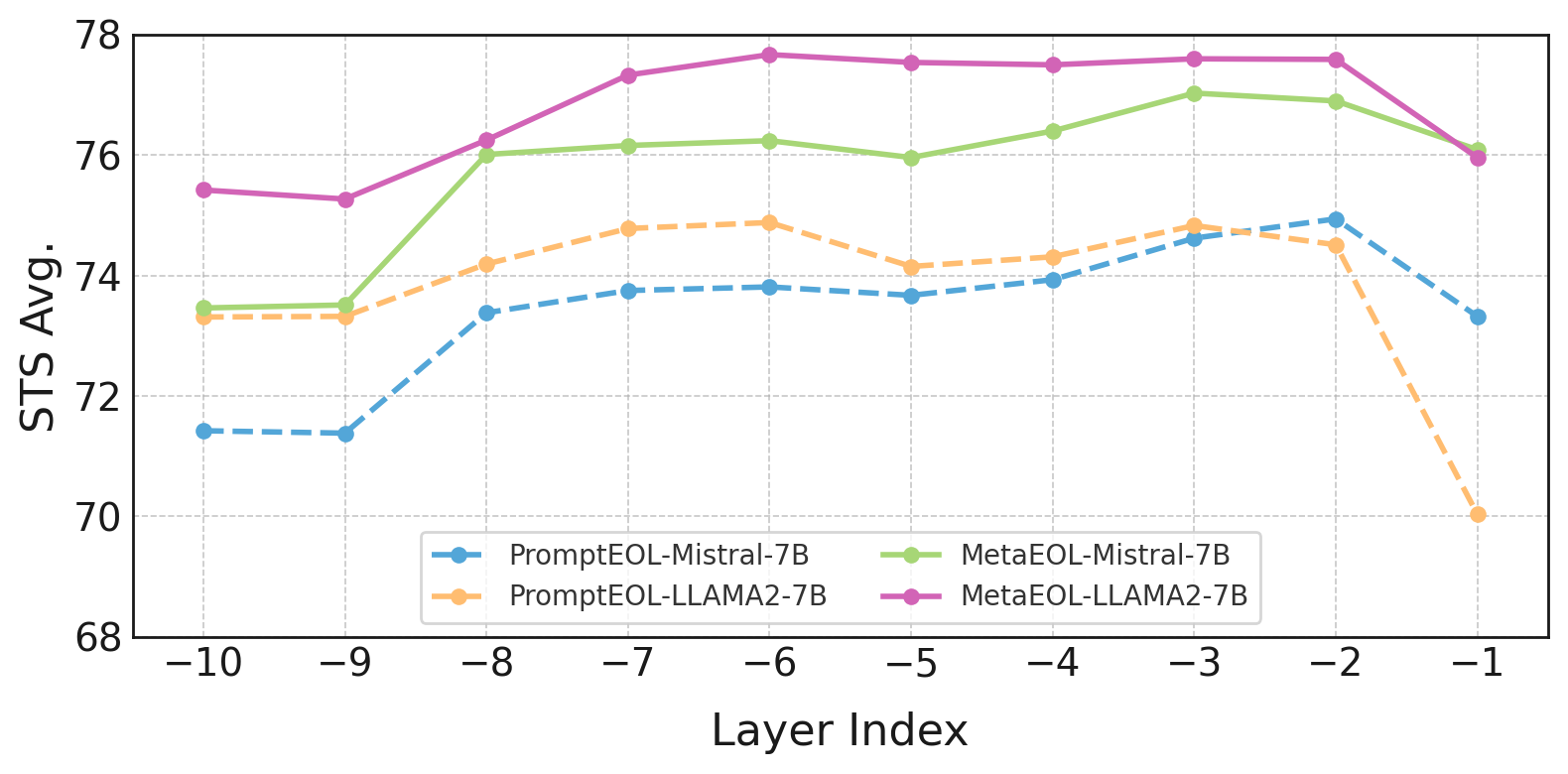}
    \caption{Influence of output layer index. STS Avg. refers to the average score of the seven STS tasks.}
    \label{fig:output_layer}
\end{figure}

We check the impact of output layers for LLAMA2 and Mistral-7B models, using PromptEOL and \ours. Figure~\ref{fig:output_layer} shows that the last layer is not always the most effective for STS tasks, which is consistent with the findings in \citet{li2024bellm}.

It is highlighted that the third-to-last layers (indexed as -3) across all four configurations perform similarly well, which suggests that this layer can be considered as a point of convergence in terms of optimal performance for these models.

\ours outperforms PromptEOL across all layers and configurations. Interestingly, PromptEOL tends to show more variability in performance across different layers compared to \ours. This suggests that the \ours approach potentially stabilizes the representational quality across layers.

\begin{table}[ht]
\centering
\small
\setlength{\tabcolsep}{5pt}
\setlength{\belowcaptionskip}{-0.5cm}
{%
\begin{tabular}{lcc}
\toprule
\textbf{Model} & \textbf{Layer Index}  & \textbf{STS Avg.}\\
\midrule
LLAMA2-7B & -1  & 75.35\\
LLAMA2-13B & -1  & 74.96\\
LLAMA2-70B & -1  & 75.41\\
\midrule
LLAMA2-7B & -3  & 77.00\\
LLAMA2-13B & -4  & 76.08\\
LLAMA2-70B & -8  & 78.06\\
\bottomrule
\end{tabular}}
\caption{Results of \ours on increasing the model size. All models are loaded with 4-bit precision. We develop a proportional layer selection strategy, leveraging the last 10\% of layers to derive sentence embeddings (specifically, the third-to-last, fourth-to-last, and eighth-to-last layers for the 7B, 13B, and 70B models, respectively), and obtain the best results with the 70B model.}
\label{tab:model_size}
\end{table}

\begin{table*}[t] 
\centering
\normalsize
\setlength{\tabcolsep}{5pt}
\setlength{\belowcaptionskip}{-0.3cm}
\resizebox{\linewidth}{!}{%
\begin{tabular}{lccccccccc}
\toprule
\textbf{Method}\quad\quad\quad\quad\quad & \quad\quad\textbf{Params}\quad\quad & \textbf{MR} & \textbf{CR} & \textbf{SUBJ} & \textbf{MPQA} & \textbf{SST} & \textbf{TREC} & \textbf{MRPC} & \textbf{Avg.}\\
\midrule
\midrule
\multicolumn{10}{l}{\it{Fine-tuning on supervised datasets}}\\
{SimCSE-RoBERTa} & 123M & 84.92 & 92.00 & 94.11 & 89.82 & 91.27 & 88.80 & 75.65 & 88.08 \\
 
ST5-Enc & 4.8B & 90.83 & 94.44 & 96.33 & 91.68 & 94.84 & 95.40 & 77.91 & 91.63 \\
\midrule
\midrule
\multicolumn{10}{l}{\it{Without fine-tuning}}\\
MRPrompt-LLAMA2 & 7B & \textbf{\textit{91.82}} & 92.88 & 97.07 & 91.60 & 96.54 & 95.80 & 74.61 & 91.47\\
CRPrompt-LLAMA2 & 7B & 91.17 & \textbf{\textit{93.27}} & 96.62 & 91.75 & 96.60 & 95.80 & 73.22 & 91.20\\
SUBJPrompt-LLAMA2 & 7B &91.88 & 93.17 & \textbf{\textit{96.96}} & 91.09 & 95.66 &  96.00 & 76.41 & 91.60\\
MPQAPrompt-LLAMA2 & 7B & 91.10 & 93.04 & 96.30 & \textbf{\textit{91.82}} &95.72 & 96.00 & 75.42 & 91.34\\
SSTPrompt-LLAMA2 & 7B &  91.82 & 92.88 & 97.07 & 91.60 & \textbf{\textit{96.54}} & 95.80 & 74.61 & 91.47 \\
TRECPrompt-LLAMA2 & 7B & 88.97 & 92.19 & 96.23 & 91.45 & 94.18 & \textbf{\textit{96.80}} & 74.72 & 90.65\\
MRPCPrompt-LLAMA2 & 7B &  90.33 & 93.32 & 96.36 & 91.45& 94.67& 96.00 & \textbf{\textit{75.13}} & 91.04\\
\multicolumn{9}{r}{\it{Avg. on task-specific prompting (i.e., diagonal):}}  & \textbf{\textit{91.76}}\\
\midrule
{PromptEOL-LLAMA2}
& 7B & 90.63 & 92.87 & 96.32 & 91.19 & 95.00 & 95.40 & 75.19 & 90.94 \\
MetaEOL-LLAMA2 (\textbf{\textit{Ours}}) & 7B & 90.93 & 93.51 & 96.12 & 91.95 &  95.77 & 97.60 & 76.81 & 91.81 \\
\bottomrule
\end{tabular}}
\caption{Results on transfer learning tasks. We design task-specific prompts for each task, denoted as \{TASK\}Prompt where \{TASK\} is a placeholder for the task's name. The corresponding task performance of each specific prompt and their average is \textbf{\textit{bold italic}}. SST and MR share the same prompt. These task-specific prompts can significantly improve the performance of the corresponding tasks compared to both PromptEOL and ST5-Enc. \ours yields superior results even without being explicitly customized for these tasks.}
\label{tab:transfer}
\end{table*}

\subsection{Scaling LLMs}
Previous study~\cite{jiang2023scaling} show scaling model sizes does not lead to performance improvement on STS tasks.
In this section, we investigate the impact of model size on the performance of \ours. For the sake of computational resources, we load models with 4-bit precision.

Informed by the insights observed from Section~\ref{sec:output_layer}, which suggested that for 7B models, the layer index -3 can be considered optimal, as evidenced by its performance in both PromptEOL and \ours. We, therefore, propose a simple proportional layer selection strategy, opting for layers -3 of 32, -4 of 40, and -8 of 80 as the output layers for the LLAMA2-7B, LLAMA2-13B, and LLAMA2-70B models respectively. This approach aligns with the model sizes, which correlates to 10\% from the final layer.

The results in Table~\ref{tab:model_size} show that using the final layer for sentence embedding generation, which is indicated by layer index -1, does not yield improved performance with increased model size. Contrastingly, the application of our proportional layer strategy reveals a different trend. Specifically, the LLAMA2-70B model, which utilizes the -8 layer, demonstrates superior performance, suggesting that larger models might benefit more significantly from selecting a proportionate layer rather than the last layer for sentence embedding. This observation could point to a potential scaling law, where larger models require a different, non-final layer to maximize performance effectively.

\subsection{Transfer Learning Tasks}
We conclude our analysis by assessing the performance of \ours on transfer learning tasks. Following prior works~\citep{gao2021simcse, ni-etal-2022-sentence}, we utilize the standard transfer learning tasks provided by SentEval. The tasks consist of MR~\citep{mr}, CR~\citep{hu2004mining_cr}, SUBJ~\citep{subj}, MPQA~\citep{wiebe2005annotating_mpqa}, SST-2~\citep{socher2013recursive_sst-2}, TREC~\citep{trec}, and MRPC~\citep{mrpc2005}.
For each task, logistic regression classifiers are trained using the created sentence embeddings as input features. The test accuracy on each task is used as the metric. Additionally, 
we include two supervised contrastive-trained models (SimCSE and ST5-Enc) for reference. Notably, ST5-Enc, a model with a 4.8B parameter count, is extensively trained on natural language inference (NLI) data and two billion question-answer pairs.

To investigate the ability of task-specific prompts to modify embedding behavior, we have crafted task prompts tailored to each SentEval task.\footnote{The details of the task prompts are in Appendix~\ref{sec:task_specific_prompts}.} As an example, for the Movie Review (MR) dataset, we designed a prompt structured as: \textit{In this task, you're given a movie review, and you need to classify its sentiment into positive or negative. For this task, this sentence: "input sentence" means in one word:"}, referred to as MRPrompt in Table~\ref{tab:transfer}. 
These task-specific prompts significantly improve the corresponding task performance, always better than PromptEOL and heavily supervised contrastive-trained ST5-Enc, verifying that LLAMA2-7B can follow the prompt to generate tailored embeddings without any training. This indicates that carefully designed prompts can effectively steer the pre-trained embeddings to align with various NLP tasks, thus providing a more resource-efficient alternative to the traditional fine-tuning paradigm. 

Moreover, although without being explicitly customized for these tasks, \ours achieves the highest average result, even outperforming heavily trained ST5-Enc. 
This suggests that the integration of the four meta-tasks in \ours can cultivate generalized embeddings that perform admirably across different tasks.

\section{Conclusion}
In this paper, we introduce \ours, a new approach for deriving high-quality sentence embeddings from LLMs without requiring any training. By leveraging a diverse set of meta-task prompts, \ours effectively captures multiple representations of sentences from distinct perspectives. We show simply averaging these meta-task derived embeddings leads to generalized general-purpose embeddings, which work remarkably well across STS datasets and transfer learning tasks.

\section*{Limitations}
We note two limitations in our work: computational overhead and restricted evaluation benchmarks. As \ours requires feeding multiple prompts to LLMs to generate several embeddings, the computational cost will be higher than that of previous methods. Our results indicate that increasing the number of tasks leads to performance improvements, but it also worsens the efficiency issue. If the number of prompts is increased, the efficiency of our approach would further decrease. Nonetheless, in contexts where sentences are consistently reused, such as when embeddings are stored for downstream classification or retrieval tasks, the issue becomes less significant.
Our evaluation is currently confined to sentence-level tasks in English only. As LLMs continue to advance, exploring the performance of \ours in multilingual contexts and its applicability to document retrieval~\cite{zhuang2024promptreps} presents an intriguing avenue for future research.

\section*{Acknowledgement}
This research was supported by the Hybrid Intelligence Center, a 10-year program funded by the Dutch Ministry of Education, Culture and Science through the Netherlands Organisation for Scientific Research, \url{https://hybrid-intelligence-centre.nl}, and project VI.Vidi.223.166 of the NWO Talent Programme which is (partly) financed by the Dutch Research Council (NWO).

\bibliography{custom}

\appendix
\onecolumn
\section{Appendix}
\label{sec:appendix}

\subsection{Instruction to Prompt ChatGPT4 for Template Generation}
We insert a blank line between paragraphs to enhance readability.
\label{sec:template_generate}
\begin{tcolorbox}
Obtaining the representation of sentences is a fundamental task in natural language processing.\\ \\
The representation can not only be used to compute the semantic similarity between different sentences but also to be directly used for downstream tasks, like Text
Categorization, Sentiment Analysis, Summarization, Style Transfer, Text Simplification, and Sentence Composition.\\ \\
A common way to obtain the representation is to use the format "This sentence "input sentence" means in one word:"" and use the hidden states of the last token as the representation of the sentence. However, we want a versatile representation that covers various aspects of the sentence by adding task instructions before the format. For instance: "In this task, you're given a review from Amazon. Your task is to generate a rating for the product on a scale of 1-5 based on the review. The rating means 1: extremely poor, 2: poor, 3: neutral, 4: good, 5: extremely good. For this task, this sentence : "input sentence" means in one word:"" is used to obtain the representation of the sentence conditioned on the given task.\\ \\
Can you help me write task instructions that can cover different aspects of the sentence such that the representation is versatile to both similarity tasks and downstream tasks?\\ \\
Please write two instructions for each of the Text Classification, Sentiment Analysis, Paraphrase Identification, and Information Extraction tasks.
\end{tcolorbox}

\subsection{Paraphrased Prompts of PromptEOL}
\label{sec:prompteol_paraphrases}
\begin{tcolorbox}
\textbf{1.} This sentence : "input sentence" can be rephrased to one word:" \\
\textbf{2.} This sentence : "input sentence" can be expressed as one word:" \\
\textbf{3.} This sentence : "input sentence" implies in one word:" \\
\textbf{4.} This sentence : "input sentence" indicates in one word:" \\
\textbf{5.} The meaning of this sentence : "input sentence" can be conveyed in another word:"\\
\textbf{6.} This sentence : "input sentence" can be restated as one word:"\\
\textbf{7.} This sentence : "input sentence" can be reformulated as one word:"
\end{tcolorbox}

\newpage

\subsection{Prompts of \ours}
\label{sec:metaeol_prompts}
\begin{tcolorbox}
\textit{\textbf{Text Classification}}\\
\textbf{General Category Identification}: In this task, you're presented with a text excerpt. Your task is to categorize the excerpt into a broad category such as 'Education', 'Technology', 'Health', 'Business', 'Environment', 'Politics', or 'Culture'. These categories help in organizing content for better accessibility and targeting. For this task, this sentence : "input sentence" should be classified under one general category in one word:"\\
\textbf{Opinion vs. Fact Discrimination}: In this task, you're given a statement and you need to determine whether it's presenting an 'Opinion' or a 'Fact'. This distinction is vital for information verification, educational purposes, and content analysis. For this task, this sentence : "input sentence" discriminates between opinion and fact in one word:"\\ \\
\textit{\textbf{Sentiment Analysis}}\\
\textbf{Product Review Rating}: In this task, you're given a review from an online platform. Your task is to generate a rating for the product based on the review on a scale of 1-5, where 1 means 'extremely negative' and 5 means 'extremely positive'. For this task, this sentence : "input sentence" reflects the sentiment in one word:"\\
\textbf{Emotion Detection}: In this task, you're reading a personal diary entry. Your task is to identify the predominant emotion expressed, such as joy, sadness, anger, fear, or love. For this task, this sentence : "input sentence" conveys the emotion in one word:"\\ \\
\textit{\textbf{Paraphrase Identification}}\\
\textbf{Similarity Check}: In this task, you're presented with two sentences. Your task is to assess whether the sentences convey the same meaning. Use 'identical', 'similar', 'different', or 'unrelated' to describe the relationship. To enhance the performance of this task, this sentence : "input sentence" means in one word:"\\
\textbf{Contextual Synonym Detection}: In this task, you're given a sentence and a phrase. Your task is to determine if the phrase can be a contextual synonym within the given sentence. Options include 'yes', 'no', or 'partially'. To enhance the performance of this task, this sentence : "input sentence" means in one word:"\\ \\
\textit{\textbf{Information Extraction}}\\
\textbf{Key Fact Identification}: In this task, you're examining a news article. Your task is to extract the most critical fact from the article. For this task, this sentence : "input sentence" encapsulates the key fact in one word:"\\
\textbf{Entity and Relation Extraction}: In this task, you're reviewing a scientific abstract. Your task is to identify the main entities (e.g., proteins, diseases) and their relations (e.g., causes, treats). For this task, this sentence : "input sentence" highlights the primary entity or relation in one word:"
\end{tcolorbox}

\newpage

\subsection{Prompts of the Sentiment Analysis Meta-Task}
\label{sec:sa_prompts}
\begin{tcolorbox}
\textit{\textbf{Sentiment Analysis Meta-Task}}\\ 
\textbf{Product Review Rating}: In this task, you're given a review from an online platform. Your task is to generate a rating for the product based on the review on a scale of 1-5, where 1 means 'extremely negative' and 5 means 'extremely positive'. For this task, this sentence : "input sentence" reflects the sentiment in one word:" \\
\textbf{Emotion Detection}: In this task, you're reading a personal diary entry. Your task is to identify the predominant emotion expressed, such as joy, sadness, anger, fear, or love. For this task, this sentence : "input sentence" conveys the emotion in one word:" \\
\textbf{Sentiment Polarity Detection}: In this task, you're analyzing customer feedback from various platforms. Your task is to identify the overall sentiment polarity of the feedback. The sentiment polarity means: 1 for very negative, 2 for negative, 3 for neutral, 4 for positive, and 5 for very positive. Based on this guidance, this sentence : "input sentence" represents in one word:"\\
\textbf{Sentiment Intensity and Emotion Detection}:
In this task, your objective is to gauge the intensity and type of emotion conveyed in a piece of text, such as a social media post or a product review. This involves not just identifying whether the sentiment is positive or negative, but also understanding the strength of that sentiment and the specific emotions involved (e.g., joy, anger, sadness, surprise). For this task, this sentence : "input sentence" conveys an emotion that is best described in one word as:"\\
\textbf{Aspect-based Sentiment Analysis}:
In this task, you're given a review of a product or service. Your task is to assess the sentiment toward specific aspects of the product or service mentioned in the review. For each mentioned aspect (e.g., quality, price, customer service), classify the sentiment as: 1 for very negative, 2 for negative, 3 for neutral, 4 for positive, and 5 for very positive. Based on this instruction, this sentence : "input sentence" signifies in one word:"
\end{tcolorbox}

\subsection{Sentiment Analysis Task Prompts with Tiny Perturbatios}
\label{sec:tiny_perturbed_prompts}
\begin{tcolorbox}
\textbf{Original}\\
In this task, you're given a review from an online platform. Your task is to generate a rating for the product based on the review on a scale of 1-5, where 1 means 'extremely negative' and 5 means 'extremely positive'. For this task, this sentence : "input sentence" reflects the sentiment in one word:"\\ \\
\textbf{Perturbed}\\
\textbf{1.} In this task, you're given a reappraisal from an online chopine. Your task is to generate a rating for the product based on the reappraisal on a scale of 1-5, where 1 think of 'extremely negative' and 5 think of 'extremely positive'. For this task, this sentence : "input sentence" reflects the sentiment in one word:"\\ 
\textbf{2.} In this task, you're given a review from an online chopine. Your task is to generate a rating for the product based on the review on a scale of 1-5, where 1 means 'extremely damaging' and 5 means 'extremely plus'. For this task, this sentence : "input sentence" reflects the sentiment in one word:"\\ 
\textbf{3.} In this job, you're given a brush up from an online platform. Your job is to generate a rating for the product based on the brush up on a scale of 1-5, where 1 means 'highly negative' and 5 means 'highly positive'. For this task, this sentence : "input sentence" reflects the sentiment in one word:"\\ 
\textbf{4.} In this task, you're reach a refresh from an online platform. Your task is to generate a rating for the product based on the refresh on a scale of 1-5, where 1 means 'highly negative' and 5 means 'highly positive'. For this task, this sentence : "input sentence" reflects the sentiment in one word:"
\end{tcolorbox}

\newpage

\subsection{Task-Specific Prompts on Transfer Tasks}
\label{sec:task_specific_prompts}
\begin{tcolorbox}
\textbf{MR/SST}\\
In this task, you're given a movie review, and you need to classify its sentiment into positive or negative. For this task, this sentence : "input sentence" means in one word:" \\ \\
\textbf{CR}\\
In this task, you're given a customer review of a product sold online, and you need to classify its sentiment into positive or negative. For this task, this sentence : "input sentence" means in one word:"\\ \\
\textbf{SUBJ}\\
In this task, you're analyzing movie reviews to determine their level of subjectivity. A subjective review is filled with personal opinions, feelings, and preferences of the reviewer, often expressing likes or dislikes and personal experiences. An objective review, on the other hand, sticks to factual information, such as plot details or actor performances, without revealing the reviewer's personal stance. For this task, this sentence : "input sentence" means in one word:"\\ \\
\textbf{MPQA}\\
In this task, you are given a description of a entity or event expressed in data such as blogs, newswire, and editorials. You need to classify its sentiment into positive or negative. For this task, this sentence : "input sentence" means in one word:"\\ \\
\textbf{TREC}\\
In this task, you are given a question. You need to detect which category better describes the question. A question belongs to the description category if it asks about description and abstract concepts. Entity questions are about entities such as animals, colors, sports, etc. Abbreviation questions ask about abbreviations and expressions abbreviated. Questions regarding human beings, description of a person, and a group or organization of persons are categorized as Human. Quantity questions are asking about numeric values and Location questions ask about locations, cities, and countries. Answer with "Description", "Entity", "Abbreviation", "Person", "Quantity", and "Location". For this task, this sentence : "input sentence" means in one word:"\\ \\
\textbf{MRPC}\\
In this task, you are given two sentences(Sentence1 and Sentence2). Answer "Yes" if these sentences are a paraphrase of one another, otherwise answer "No". For this task, this sentence : "input sentence" means in one word:"
\end{tcolorbox}

\end{document}